\documentclass[letterpaper]{article} % DO NOT CHANGE THIS
\usepackage{aaai24}  % DO NOT CHANGE THIS
\usepackage{times}  % DO NOT CHANGE THIS
\usepackage{helvet}  % DO NOT CHANGE THIS
\usepackage{courier}  % DO NOT CHANGE THIS
\usepackage[hyphens]{url}  % DO NOT CHANGE THIS
\usepackage{graphicx} % DO NOT CHANGE THIS
\usepackage{natbib}  % DO NOT CHANGE THIS AND DO NOT ADD ANY OPTIONS TO IT
\usepackage{caption} % DO NOT CHANGE THIS AND DO NOT ADD ANY OPTIONS TO IT
\usepackage{algorithm}
\usepackage{algorithmic}
\usepackage{multirow}
\usepackage{enumitem}
\usepackage[]{mdframed}

\usepackage{newfloat}
\usepackage{listings}
\DeclareCaptionStyle{ruled}{labelfont=normalfont,labelsep=colon,strut=off} % DO NOT CHANGE THIS
\floatstyle{ruled}
\newfloat{listing}{tb}{lst}{}
\floatname{listing}{Listing}
\usepackage{amsmath}
\usepackage{amsfonts}
\usepackage{multirow}
\usepackage{enumitem}
\usepackage[]{mdframed}
\usepackage{etoolbox}
\makeatletter
\patchcmd{\@verbatim}
  {\verbatim@font}
  {\verbatim@font\tiny}
  {}{}
\makeatother

\newcommand{\G}{\mathcal{G}}
\newcommand{\Loss}{\mathcal{L}}

\title{Enhancing Student Performance Prediction on Learnersourced Questions with SGNN-LLM Synergy}
\author {
    Lin Ni\textsuperscript{\rm 1,2 \textdagger},
    Sijie Wang\textsuperscript{\rm 2},
    Zeyu Zhang\textsuperscript{\rm 1},
    Xiaoxuan Li\textsuperscript{\rm 2},
    Xianda Zheng\textsuperscript{\rm 2},
    Paul Denny\textsuperscript{\rm 2},
    Jiamou Liu\textsuperscript{\rm 2 \textdaggerdbl}
}
\affiliations {
    \textsuperscript{\rm 1}Huazhong Agricultural University, Wuhan 430070, China\\
    \textsuperscript{\rm 2}School of Computer Science, The University of Auckland, Auckland 1010, New Zealand\\
    \textdagger l.ni@auckland.ac.nz, \textdaggerdbl jiamou.liu@auckland.ac.nz
}

\begin{document}

\maketitle

\begin{abstract}
Learnersourcing offers great potential for scalable education through student content creation. However, predicting student performance on learnersourced questions, which is essential for personalizing the learning experience, is challenging due to the inherent noise in student-generated data. Moreover, while conventional graph-based methods can capture the complex network of student and question interactions, they often fall short under cold start conditions where limited student engagement with questions yields sparse data.  To address both challenges, we introduce an innovative strategy that synergizes the potential of integrating Signed Graph Neural Networks (SGNNs) and Large Language Model (LLM) embeddings. Our methodology employs a signed bipartite graph to comprehensively model student answers, complemented by a contrastive learning framework that enhances noise resilience. Furthermore, LLM's contribution lies in generating foundational question embeddings, proving especially advantageous in addressing cold start scenarios characterized by limited graph data. 
Validation across five real-world datasets sourced from the PeerWise platform underscores our approach's effectiveness. Our method outperforms baselines, showcasing enhanced predictive accuracy and robustness. 
% Our implementation is accessible in PyTorch\footnote{https://anonymous.4open.science/r/Education\_system-DD40}, offering a practical avenue for implementation and further exploration.
\end{abstract}

% \keywords{learnersourcing platform, student performance prediction, cold start problem, large language model, graph neural networks, signed bipartite graph, contrastive learning.}

\section{Introduction}\label{sec:intro}
The rapid growth of online education in recent years has resulted in considerable interest in new approaches for delivering effective learning at scale \cite{xianghan2022virtually}.
A significant challenge for educators adopting new online learning platforms is the need to develop new content \cite{mulryan2010teaching}. 
Educators face the task of creating extensive repositories of items for supporting personalized learning and exploring innovative methods to immerse students in such material. Learnersourcing, a relatively new approach, has emerged as a promising technique to address the challenges of content development and student engagement in online learning \cite{khosravi2021charting, denny2022robosourcing}. It involves leveraging the collective knowledge, skills, and contributions of learners to create educational resources. In general, accurate prediction of student performance \cite{li2020peer,li2022multi}, such as future grades on assignments and exams, is immensely valuable for informing instructional strategies and  enabling personalized support.  Within learnersourcing contexts, performance prediction on individual items can support the recommendation of new items from large repositories to match the current ability level of students \cite{paramythis2003adaptive}.

Multiple-choice questions (MCQs) play a pivotal role in learnersourcing platforms like RiPPLE \cite{khosravi2019ripple}, Quizzical \cite{riggs2020positive}, UpGrade \cite{wang2019upgrade}, and PeerWise \cite{denny2008peerwise}. Predicting student performance on MCQs is a crucial research challenge \cite{abdi2021modelling}. MCQs, comprising a stem and multiple answer options, assess a wide range of cognitive skills, from factual recall to higher-order thinking (e.g., analysis, synthesis, and evaluation). They offer efficient grading and valuable insights into subject comprehension.

As learners engage in creating and responding to MCQs on platforms like PeerWise, a graph emerges, with learners and MCQs as nodes and their interactions as edges. GNNs provide a powerful means to capture these intricate learner-to-question, learner-to-learner, and question-to-question relationships \cite{zhou2020graph}. GNNs excel at extracting nuanced features from graph topology, offering insights beyond traditional methods.
However, while GNNs hold promise for predictive analytics in learnersourcing platforms, applying them to anticipate student performance on MCQs is relatively unexplored. A key challenge lies in modeling both students' responses and their correctness (i.e., correct or erroneous answers) within noisy data. Noise can stem from inadvertent errors during question creation or participation dynamics discouraging accurate responses. 
Furthermore, learnersourcing repositories constantly evolve, presenting a cold start dilemma. Newly created MCQs with limited responses lack substantial interaction data, hampering effective GNN utilization and predictive accuracy.

In summary, our study addresses two crucial challenges for predicting student performance on MCQs in learnersourcing platforms:

\begin{itemize}[leftmargin=*]
   \item Modeling complex relationships involving students and MCQs, accommodating both accurate and erroneous responses within noisy data.
   \item Tackling the cold start problem inherent to questions with low answer rates, where insufficient graph data undermines predictive accuracy.
\end{itemize}

\begin{figure}[H]
    \centering
    \includegraphics[width=0.95\columnwidth]{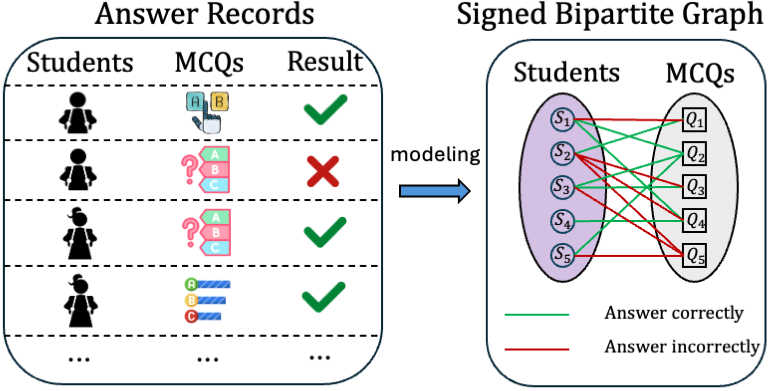}
    \caption{A scenario for the signed bipartite graph}
    \label{fig:Example}
\end{figure}

To address these challenges, we propose an innovative technique that leverages the potential of GNNs for application to the unique dynamics present in learnersourcing repositories. Our approach begins by adopting a signed bipartite graph to model students' response accuracy, as illustrated in Figure~\ref{fig:Example}. This graph structure segregates students and questions into separate sets of nodes. Positive edges denote correct student answers, while negative edges indicate incorrect responses. To counter the impact of noise, we introduce a novel contrastive learning framework tailored for signed bipartite graphs. This approach draws from the emerging paradigm of graph contrastive learning \cite{shuai2022review,zhu2021graph,zhu2020deep,you2020graph}, renowned for its ability to resist noise by contrasting similar and dissimilar instances. Our choice of contrastive learning not only enhances model resilience but also addresses noise interference within the data.

Furthermore, we pioneer the integration of NLP technology to extract intrinsic knowledge essential for accurate MCQ responses, offering a solution to the cold start problem. Although previous student performance prediction research seldom leveraged NLP advancements, recent studies underscore the potential of integrating advanced NLP in education to unravel question intricacies \cite{ni2022deepqr}. We posit that MCQ response accuracy hinges on students' comprehension of underlying knowledge within questions. Leveraging the gpt-3.5-turbo-0613 LLM\footnote{https://platform.openai.com/docs/models/continuous-model-upgrades}, we capture nuances within MCQs, extracting and weighing inherent knowledge points. By fusing this semantic information with structural embeddings derived from the graph, our model gains a holistic view for enhanced student performance prediction, particularly in cold start scenarios. Our main contributions include:

\begin{enumerate}[leftmargin=*]
\item Formulating the student performance prediction problem as a sign prediction on a signed bipartite graph.
\item Exploiting contrastive learning to enhance the robustness of representations for students and questions.
\item Employing an LLM to extract semantic embeddings of key knowledge points in questions, and integrating this information with structural embeddings learned from a graph to address the cold start problem.
\item Validating the efficacy of our proposed model through experiments conducted on five real-world datasets sourced from a prominent learnersourcing platform, achieving the highest f1 value of 0.908.
\end{enumerate}

\section{Related Work}
We investigate four domains: student academic performance prediction, the cold start problem, GNNs, and LLMs.

\textbf{Performance Prediction}: Significant research has been devoted to predicting student performance on online learning platforms, with the aim of personalizing and improving the learning experience \cite{wang2023unified}. This body of work primarily encompasses static and sequential models \cite{li2020peer, thaker2019comprehension}. Static models, exemplified by Logistic Regression, leverage historical data such as student scores and learning activities to forecast future performance \cite{jiang2014predicting, wei2020predicting, daud2017predicting}. In contrast, sequential models like Knowledge Tracing and its variants explore the sequential relationships within learning materials, allowing dynamic knowledge updates \cite{piech2015deep, nakagawa2019graph}. Knowledge Tracing, a subset of educational data mining and machine learning, aims to model and predict learners' knowledge or skill acquisition over time \cite{lyu2022deep}. However, our task presents unique challenges. Unlike many previous approaches, we lack access to student profiles and additional contextual information. Furthermore, we do not possess information about the temporal sequence of learner responses. Consequently, our research is focused on exploring deep learning solutions that rely solely on question and learner-question interaction data to enhance student academic performance prediction.

\textbf{Cold Start Problem}: Addressing the cold start challenge is a fundamental concern in representation learning, particularly in machine learning scenarios in e-learning environments where minimal prior information is available about the target data \cite{liu2022review}. Researchers have explored diverse strategies to tackle this issue, tailoring their approaches to the specific nature of the problem, available resources, and data characteristics. One common strategy is transfer learning, which entails pre-training a model on a related task with a substantial dataset and then fine-tuning it for the target task with limited data \cite{pang2022pnmta}. This method leverages prior knowledge gained during pre-training to mitigate cold start challenges. In cases where data is scarce, knowledge-based systems employ explicit rules or domain knowledge to provide recommendations or predictions, bypassing the need for extensive historical data \cite{du2022metakg}. Recommendation systems often turn to content-based filtering, relying on item and user attributes or features for making recommendations \cite{schein2002methods, safarov2023deep}. This approach proves effective, particularly in scenarios with little or no historical user-item interaction data, by suggesting items based on their inherent characteristics. In alignment with previous studies that have leveraged ontologies as supplementary data to mitigate the cold start problem \cite{sun2017towards, jeevamol2021ontology}, our approach capitalizes on the inherent semantic richness of MCQs sourced from learnersourcing platforms. These MCQs inherently contain vital knowledge points essential for predictive tasks. To effectively address the challenges posed by the cold start problem, should this be we incorporate embedding information from LLMs that encapsulate both common sense and domain-specific knowledge. This integration serves to enhance the representation of MCQs in our model.

\textbf{Signed Graph Representation Learning Methods}: Predicting student performance involves the unique task of sign prediction in a signed bipartite graph, where positive edges signify correct answers and negative edges denote incorrect ones \cite{derr2018signed,zhang2023rsgnn}. Traditional methods like SIDE \cite{kim2018side} and SGDN \cite{jung2020signed} leveraged random walks, while neural-based approaches like SGCN \cite{derr2018signed} extended Graph Convolutional Networks (GCN) \cite{kipf2016semi} to balance theory-guided sign prediction. However, adapting these methods to signed bipartite graphs is non-trivial. Graph Contrastive Learning techniques, inspired by successes in image processing, aim to generate stable node representations under perturbations. Deep Graph Infomax (DGI) \cite{velickovic2019deep} maximizes the mutual information between global graph and local node representations. GraphCL \cite{you2020graph} introduces varied graph augmentations, while Graph Contrastive Augmentation (GCA) employs diverse views for agreement of node representations. Most of these models are designed for unipartite graphs, with Signed Graph Contrastive Learning (SGCL) \cite{shu2021sgcl} adapting to signed unipartite graphs. However, Signed Bipartite graph Contrastive Learning (SBCL) \cite{zhang2023contrastive} is the only model combining both signed and bipartite graph aspects, albeit lacking consideration for node features. Notably, these methods primarily rely on large-scale node interaction data to extract structural information for representation, which may limit their effectiveness in addressing the cold start problem.

\textbf{LLM in NLP}: Recent years have witnessed significant advancements in LLMs in natural language processing (NLP). Models like BERT~\cite{devlin2018bert} and T5~\cite{raffel2020exploring} excel in various NLP tasks by pre-training on extensive text data, followed by fine-tuning. However, fine-tuning often demands a substantial amount of task-specific data and introduces complexity. A groundbreaking development came with GPT-3~\cite{brown2020language}, a massive autoregressive language model with 175B parameters. It introduced In-Context Learning, allowing it to perform tasks without extensive fine-tuning or model updates. This innovation led to Instruction-Finetuned Language Models, exemplified by GPT-3.5~\cite{ye2023comprehensive}, which efficiently adapts to specific tasks using task-specific instructions. This approach overcomes previous limitations, making LLMs a cornerstone in NLP.

\section{Problem Definition}
In this study, we model the student performance prediction problem on a learnersourcing platform as the edge sign prediction problem on a signed bipartite graph, as illustrated in Figure \ref{fig:Example}. We define our problem within the context of a bipartite graph $\mathcal{G} = (\mathcal{U}, \mathcal{V}, \mathcal{E})$, where $\mathcal{U} = \{u_1, u_2, ..., u_{|\mathcal{U}|}\}$ and $\mathcal{V} = \{v_1, v_2, ..., v_{|\mathcal{V}|}\}$ represent the nodes for students and questions, respectively. $|\mathcal{U}|$ and $|\mathcal{V}|$ denote the total number of students and questions, where $\mathcal{U} \cap \mathcal{V} = \emptyset$. The edge set $\mathcal{E} \subset \mathcal{U} \times \mathcal{V}$ signifies the links between students and questions and is further partitioned into positive and negative edge sets, $\mathcal{E}^+$ and $\mathcal{E}^-$, respectively, with $\mathcal{E} = \mathcal{E}^+ \cup \mathcal{E}^-$ and $\mathcal{E}^+ \cap \mathcal{E}^- = \emptyset$. Given this bipartite graph and any pair of nodes $u_i \in \mathcal{U}$, $v_j \in \mathcal{V}$ and the sign of the edge $e_{ij} \in \mathcal{E}$ is unknown, our goal is to leverage GNNs to learn the embeddings $z_{u_i} \in \mathbb{R}^d$ and $z_{v_j} \in \mathbb{R}^d$ of node $u_i$ and $v_j$, respectively. Subsequently, we seek to find a mapping  $f\colon (z_{u_i}, z_{v_j}) \mapsto \{-1, +1\}$ to determine the sign of the edge, thus producing our desired result.

One significant challenge in this context is the cold start problem for newly generated questions. In early-stage situations, a limited number of learners may have attempted the question, and consequently, the bipartite graph is sparsely connected. The edges connecting students to these questions are uncertain or unknown, making the prediction of student performance particularly challenging.

\section{Proposed Method}
In this section, we introduce the \underline{L}arge \underline{L}anguage \underline{M}odel enhanced \underline{S}igned \underline{B}ipartite graph \underline{C}ontrastive \underline{L}earning (\textbf{LLM-SBCL}) model, a framework designed to learn representations for users and questions while facilitated by LLM.

\begin{figure}[htb]
    \centering
    \includegraphics[width=\columnwidth]{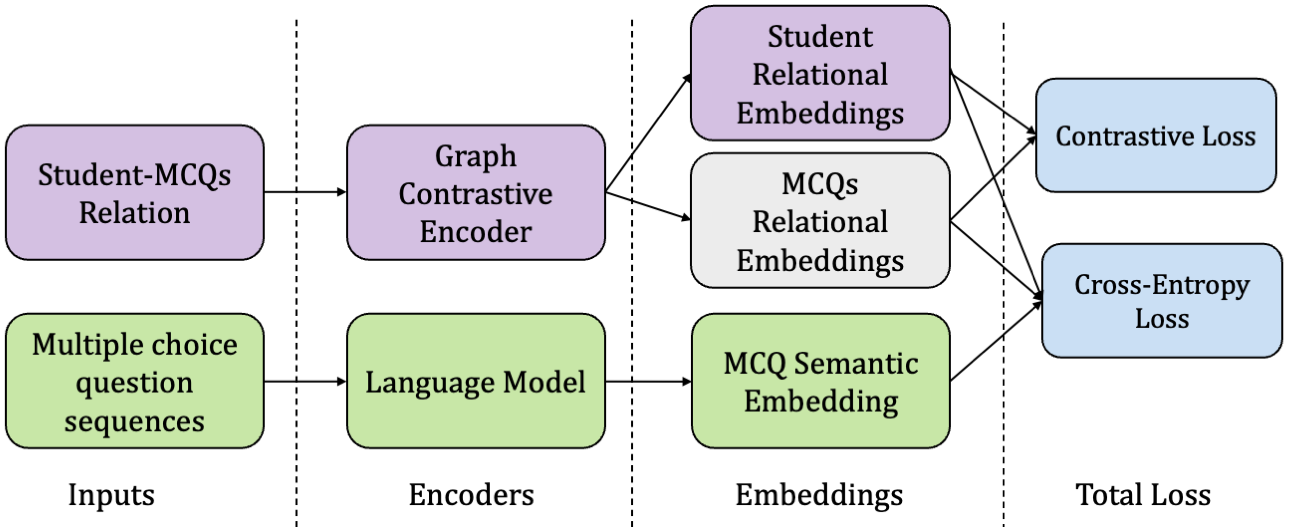}
    \caption{The Framework of LLM-SBCL model}
    \label{fig:Framework}
\end{figure}

Our framework, as depicted in Figure \ref{fig:Framework}, capitalizes on student-MCQs relationships in learnersourcing platforms using SBCL. It employs separate encoders for students and MCQs, with a graph contrastive learning model to handle dataset noise. To address cold start challenges, we enhance question embeddings with semantic knowledge points extracted by LLM. This supplementary data augments the question structure embeddings to compute cross-entropy loss. In the following sections, we provide a detailed breakdown of our model's architecture.

\subsection{Signed Bipartite Graph Contrastive Learning}
\begin{figure*}[htb]
    \centering
    \includegraphics[width=1.9\columnwidth]{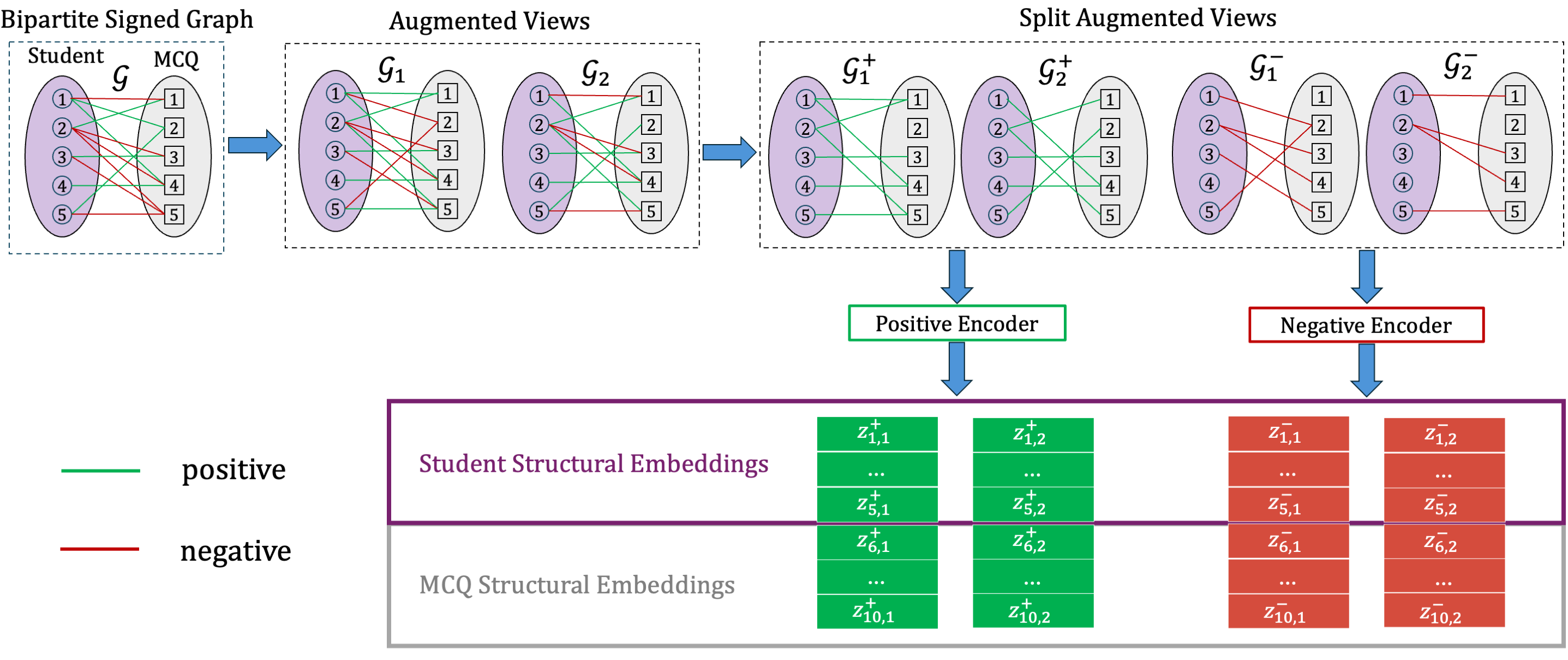}
    \caption{The framework of SBCL}
    \label{fig:graph_encoder}
\end{figure*}

Fig.~\ref{fig:graph_encoder} illustrates SBCL, which includes graph augmentation and graph encoding processes. During the graph augmentation phase, we use stochastic graph augmentation to generate two augmented graphs. Each graph is then bifurcated according to edge signs, resulting in a positive and a negative graph. These are further processed using GNNs to learn the structural embeddings for node representations.

\textbf{Graph Augmentation}: To enhance the robustness and generalization of learned node representations in the context of learnersourcing platforms with noisy data, we utilize stochastic perturbation for graph augmentation. We generate two distinct graphs from the original one by randomly flipping edge signs, transforming a positive edge into a negative one, and vice versa, with a probability $p$.

\textbf{Graph Encoder}: Following data augmentation, we acquire two augmented graphs, denoted as $\G_1$ and $\G_2$, as illustrated in Fig \ref{fig:graph_encoder}. Considering the distinct semantic attributes of positive and negative edges—positive edges representing correct answers by users and negative edges indicating incorrect answers—we employ separate GNN encoders for each edge type. This dual-GNN approach aligns with our contrastive objective, which we will delve into in the next subsection. Consequently, we partition each augmented graph into two sub-graphs exclusively featuring positive and negative edges, denoted as positive graph $\G_m^+$ and negative graph $\G_m^-$, where $m\in \{1, 2\}$. To facilitate this encoding, we adopt the Positive and Negative Graph Attention Network (GAT) \cite{velivckovic2017graph}, consistent with the design in \cite{zhang2023contrastive}. 
Formally, the representation for the $i$-th node, denoted as $z_i$, is computed as follows:

\begin{equation}\footnotesize
    z_{i,m}^{(l+1),\sigma} = \mathrm{PReLU}(\mathrm{GNN}^{\sigma}_{m}(z_{i,m}^{(l), \sigma}, \G_{m}^{\sigma}))
\end{equation}

\begin{equation}\footnotesize
\label{eq:final_z}
    z_{i} = \left[z_{i,1}^{(L), +} \| z_{i,1}^{(L),-} \| z_{i,2}^{(L),+} \| z_{i,2}^{(L),-}\right]W
\end{equation}
where $\sigma \in \{+,-\}$, $m$ refers to the $m$-th augmented graph, $L$ denotes the number of GNN layers and $l$ denotes the $l$-th layer. $z_{i,m}^{(0),\sigma}$ denotes the input feature vector of the $i$-th node. $W$ is a learn-able transformation matrix.

\begin{figure}[htb]
    \centering
    \includegraphics[width=0.825\columnwidth]{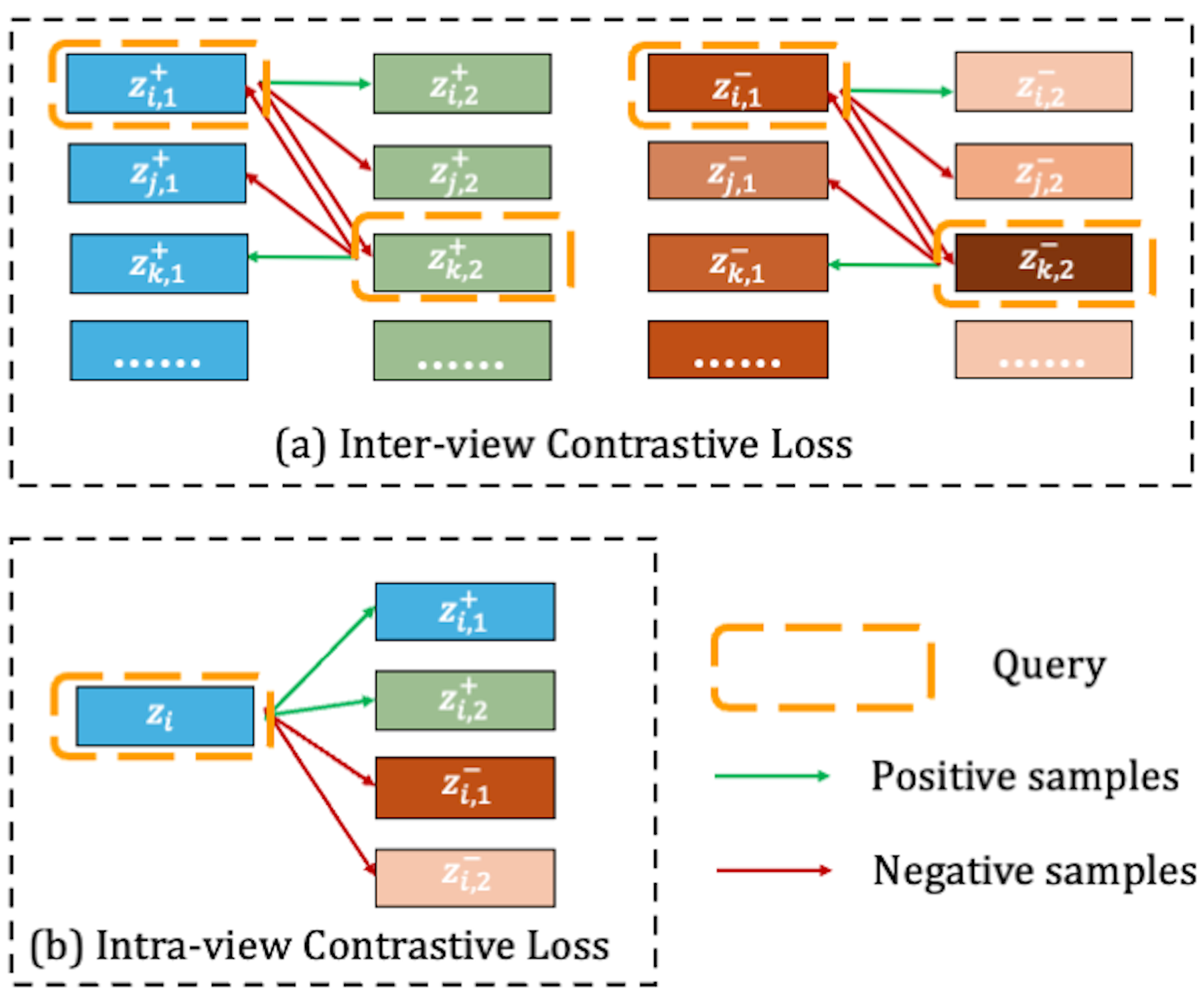}
    \caption{Inter-view and Intra-view contrastive loss}
    \label{fig:contrastive_loss}
\end{figure}

\textbf{Contrastive Objective}: Noise is commonly expected in a learnersourcing platform where the majority of content is student-generated. Variations in course rules for question creation and answering, along with student behaviors, can introduce considerable noise. For instance, when the motivation for answering a question is low (e.g., performance on such platform does not impact final scores), students might guess answers randomly. Likewise, students may unintentionally create incorrect questions. We categorize these instances as noise within our dataset. To address this, we employ a contrastive objective for robust signed bipartite graph representation learning, encompassing two losses: (a) \emph{Inter-view contrastive loss} and (b) \emph{Intra-view contrastive loss}, where (a) is for nodes from the same encoder while (b) is for those from different encoders as depicted in Fig \ref{fig:contrastive_loss}.

We utilize InfoNCE loss \cite{sohn2016improved,oord2018representation} to define our inter-view contrastive loss for positive augmented graphs as follows:

\begin{equation} \footnotesize
\Loss_{\text {inter}}^{+}=-\frac{1}{I} \sum_{i=1}^I \log \frac{\exp \left(\operatorname{sim}\left(z_{i, m}^{+}, z_{i, m^{\prime}}^{+}\right) / \tau\right)}{\sum_{j=1, j \neq i}^I \exp \left(\operatorname{sim}\left(z_{i, m}^{+}, z_{j, m^{\prime}}^{+}\right) / \tau\right)}
\end{equation}

where $I$ is the number of nodes in a mini-batch, $z^+_{i,m}$ represents the representation of node $i$ in the $m$-th augmented positive graph, $sim(\cdot,\cdot)$ represents the similarity function between the two representations (e.g., cosine similarity) and $\tau$ denotes the temperature parameter. The perspective-specific contrastive loss for negative graphs is similar.

The intra-view contrastive loss is defined as:
\begin{equation} \footnotesize
\Loss_{\text {intra}}=-\frac{1}{I} \sum_{i=1}^I \log \frac{\sum_{m=1}^M \exp \left(\operatorname{sim}\left(z_i, z_{i, m}^{+}\right) / \tau\right)}{\sum_{m=1}^M \exp \left(\operatorname{sim}\left(z_i, z_{i, m}^{-}\right) / \tau\right)}
\end{equation}
where $M$ denotes the number of augmented graphs, which is equal to 4 in our paper. The Combined contrastive loss is
\begin{equation}
\footnotesize
\Loss_{C L}=(1-\alpha) \cdot (\Loss_{\text {inter}}^{+} + \Loss_{\text {inter}}^{-})+\alpha \cdot \Loss_{\text {intra}}
\end{equation}
where $\alpha$ is the weight coefficient that controls the significance between two losses.

\subsection{Incorporating Semantic Embedding into SBCL}
In our LLM-SBCL framework (Fig \ref{fig:Framework}), we employ a multimodal approach, utilizing GNNs and NLP Models, to process User-Question Relationship and MCQ Textual Information respectively. We recognize that infusing semantic context into MCQs has great potential for uncovering the underlying knowledge, which are crucial for the prediction task.

\textbf{Extracting Semantic Embeddings via LLM}: To include semantic information, we utilize a LLM to extract knowledge from MCQs. We combine these semantic representations with relational embeddings from the Graph Encoder, training them with cross-entropy loss $\Loss_{\mathrm{CE}}$ for a more comprehensive representation.

\begin{figure}
    \centering
    \includegraphics[width=\columnwidth]{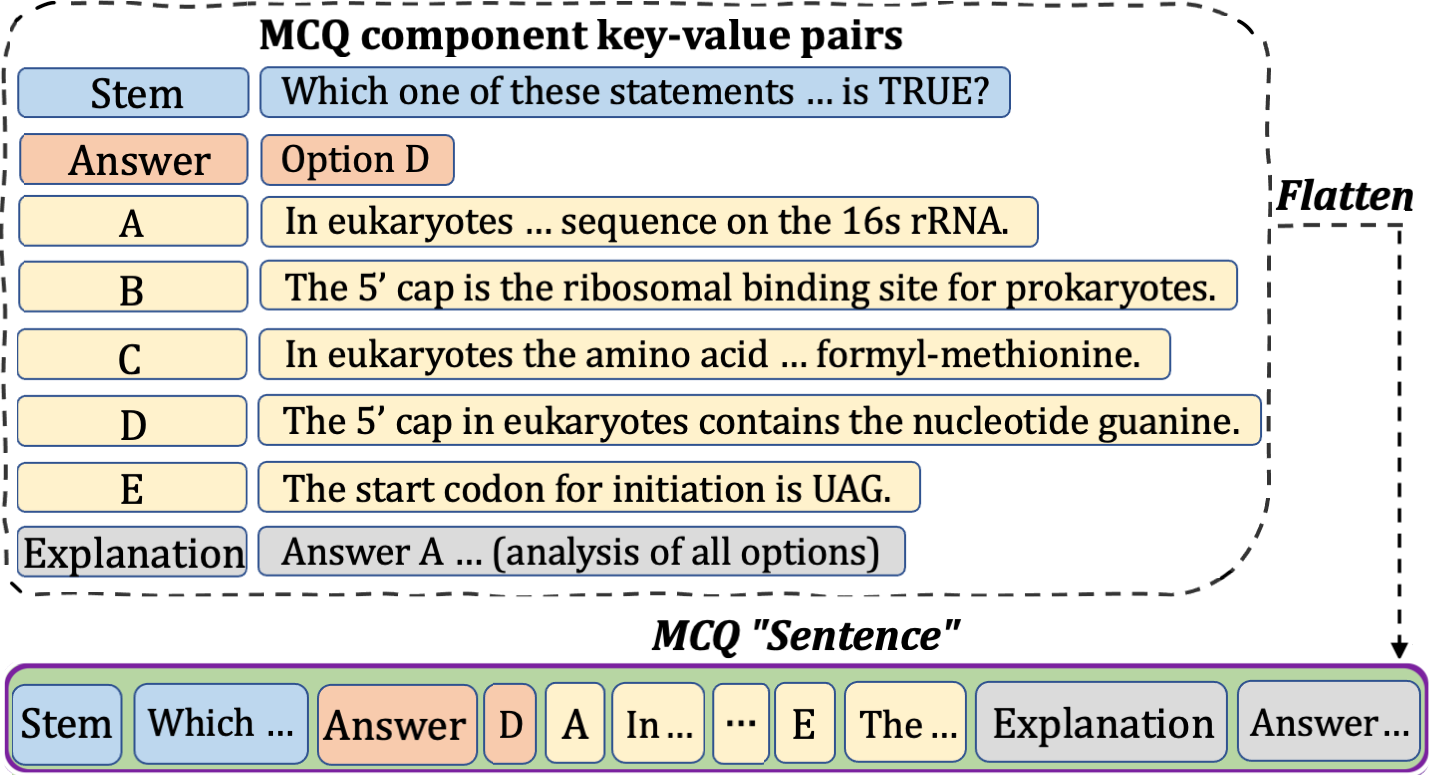}
    \caption{Flatten MCQ component key-value pairs into a sequence to form an MCQ ``sentence''}
    \label{fig:Mcq_new}
\end{figure}

Our approach involves transforming MCQs into sentence format by flattening key-value pairs (Fig \ref{fig:Mcq_new}). This results in input data, denoted as $mcq_j$, comprising content from Stem, Answer, Options, and Explanation: $mcq_j = \{S_k, S_v, A_k, A_v, O_{A,k}, O_{A,v}, \ldots, O_{E,k}, O_{E,v}, E_k, E_v\}$.

Next, we employ LLM to extract $n$ knowledge point terms $t_i$ and their corresponding weights $h_i$ for each MCQ. GloVe word embeddings \cite{pennington2014glove} provide word embeddings $emb_{i,k}$ for words in $t_i$. The average of these embeddings represents a single knowledge point $kp_i$. We compute the weighted average of all knowledge points to obtain the Question Semantic Embedding $w_j$.

\begin{equation} \footnotesize
w_j = \frac{{\sum_{i=1}^{n} h_i \cdot kp_i}}{{\sum_{i=1}^{n} h_i}}, \quad kp_i = \frac{{\sum_{k=1}^{m} emb_{i,k}}}{{m}}
\end{equation}

\textbf{NLP Large Language Model}: We employ gpt-3.5-turbo-0613 \footnote{https://platform.openai.com/docs/models/continuous-model-upgrades} as our LLM engine, capable of retrieving $n$ pairs of $t_i$ and $h_i$ with a single prompt API request.

\begin{mdframed}[skipabove=6pt]
\footnotesize
\begin{itemize}[leftmargin=*]
\item[]
\begin{footnotesize}
\item \textbf{Task Objective}: Your task is to extract key knowledge points from an MCQ created by a first-year student who studies biology in college. The question consists of a stem, up to five options, an answer, and an explanation provided by the student author.

\item \textbf{Special Requirements}: Please provide your response in JSON format with the following keys and format: 
\begin{verbatim}
{"Keywords": [{"keyword": "keyword_name", "percentage": 0}]}
\end{verbatim}

1. Each \texttt{"keyword\_name"} should be a word or a short term with less than five words.

2. The percentages of all keywords should add up to 1.

3. Only include the top five keywords to avoid excessive keyword extraction.

\item \textbf{In Context Learning}: Here's an example response format:
\begin{verbatim}
{"Keywords": [
  {"keyword": "Menstrual cycle", "percentage": 0.8}, 
  {"keyword": "Luteal phase", "percentage": 0.2 }
]}
\end{verbatim}

\item \textbf{Context}:
Here's the MCQ from the student author:

Question stem: Jamie has a Atherosclerosis, what does this imply? 

[A]: She constantly needs to visit the ladies room to pass urine. 

[B]: She has impaired vision. 

[C]: She has lost a lot of weight recently due to no apparent reason. (She hasn't suddenly turned paleo). 

[D]: Her arterial walls have thickened making her more likely to have a heart attack. 

Question answer: D

Explanation: Atherosclerosis is the thickening of the artery walls (generally due to poor diet and lack of exercise). This can lead to heat attacks and strokes, as when pressure builds up in the arteries, arteries are more likely to burst. In a normal person the walls of arteries are more likely to stretch and allow for the fluctuations of pressure in the blood. 

\item \textbf{LLM Response}:
\begin{verbatim}
[{"keyword": "Atherosclerosis", "percentage": 0.44},
{"keyword": "Arterial walls", "percentage": 0.22},
{"keyword": "Thickening", "percentage": 0.11},
{"keyword": "Heart attack", "percentage": 0.11},
{"keyword": "Strokes", "percentage": 0.11}]
\end{verbatim}
\end{footnotesize}
\end{itemize}
\end{mdframed}

The provided prompt encompasses the task objective, requirements, in-context learning format, and context, ensuring consistent and aligned responses. This well-crafted prompt guided the LLM to analyse the specific knowledge topics related to the MCQs. As a result, the Question Semantic Embedding derived from this prompt becomes more meaningful, thereby enhancing the accuracy of student performance prediction.

\textbf{Semantic Embedding Integration}: LLM-SBCL enhances SBCL with semantic question embeddings. Instead of directly predicting the edge sign between student $u_i$ and question $v_j$, we also include semantic embeddings $w_j$ for question representations.

This results in $d$-dimensional embeddings for $u_i$ and $v_j$ and an additional $d$-dimensional embedding for $w_j$. To maintain uniform dimensions, we transform $v_j$ and $w_j$ into a unified $d$-dimensional representation $q_{v_j}$ using a learnable matrix $W_q$. The transformed question embeddings, $q_{v_j}$, are concatenated with $u_i$ and processed through a single-layer MLP for edge sign prediction:

\begin{equation} \footnotesize
q_{v_j} = (v_j \| w_j) W_q, \quad y_{\mathrm{pred}}=\operatorname{MLP}\left(u_i \| q_{v_j}\right)
\end{equation}

Use the predicted $y_{pred}$ in the cross-entropy loss function:

\begin{equation} \footnotesize
\Loss_{\mathrm{CE}}=-y \cdot \log y_{\mathrm{pred}}+(1-y) \cdot \log \left(1-y_{\mathrm{pred}}\right)
\end{equation}

where $y$ is the ground truth from $\{-1,1\}$ to $\{0,1\}$.

The final joint loss of cross-entropy and contrastive loss:

\begin{equation} \footnotesize
\Loss=\Loss_{\mathrm{CE}}+\beta \cdot \Loss_{C L}
\end{equation}

Here, $\beta$ controls the contribution of the contrastive loss.

\section{Experiments}
In this section, we evaluate the LLM-SBCL model's performance on five real-world datasets: biology, law, cardiff20102, sydney19351, and sydney23146. We introduce the datasets and compare our model against state-of-the-art baselines. We provide experimental details and present results from both baseline and our models.

\subsection{Dataset Background and Introduction}
PeerWise is a platform that allows students to share their knowledge with each other. Teachers can create courses and manage student access to them. MCQs are a common type of assessment format that presents a question or statement followed by a set of predetermined options or choices. Within these courses, students can create and explain their own MCQs related to the course material. They can also answer and discuss questions posed by their peers. This allows for a collaborative learning experience where students can learn from each other and deepen their understanding of the course material.

% We use five real-world datasets from PeerWise: biology and law from the University of Auckland, cardiff20102 from Cardiff University School of Medicine with course id 20102, sydney19351 and sydney23146 are from two University of Sydney biochemistry courses with the highest number of answering records (i.e. links). Detailed summaries and balance information for these datasets are provided in Table \ref{tab:datainfo} and Table \ref{tab:databalance}.

We use five real-world datasets from PeerWise: biology and law from the University of Auckland, cardiff20102 from Cardiff University School of Medicine with course id 20102, sydney19351 and sydney23146 are from two University of Sydney biochemistry courses with the highest number of answering records (i.e. links). Detailed summaries for these datasets are provided in Table \ref{tab:data_info}.

\begin{table}[ht]
\centering
\begin{footnotesize}
\begin{tabular}{lccccc}
\hline
             & Biology & Law   & Cardiff & Sydney1 & Sydney2 \\ \hline
Students     & 761     & 528   & 383     & 382     & 198     \\
MCQs         & 380     & 5600  & 1171    & 457     & 748     \\
Links        & 76613   & 88563 & 64524   & 24032   & 24050   \\
$Link_{P\%}$  & 0.665   & 0.931 & 0.600   & 0.531   & 0.706   \\
$Link_{N\%}$  & 0.335   & 0.069 & 0.400   & 0.469   & 0.294   \\ \hline
\end{tabular}
\end{footnotesize}
\caption{Statistics on Signed Bipartite Networks.}
\label{tab:data_info}
\end{table}

Since each course is independent of the others, we construct a signed bipartite graph for each dataset using the answering records. For a student $u_i \in \mathcal{U}$ answering a question $v_j \in \mathcal{V}$, we establish an edge between $u_i$ and $v_j$ with its sign being +1 if the answer of $u_i$ matches the correct answers for $v_j$'s and -1 otherwise. We observe that the overall correctness for the questions exceeds 50\% for each course, with most courses ranging from 60\% to 70\%. However, there are exceptions such as Law, where the overall correctness reaches 93\%. The inherent nature of such online peer-practice platforms suggests an innate data imbalance issue, which is handled by class weights in the experiment.

\subsection{Experimental Settings}

We compared our model against state-of-the-art methods, including Random Embedding, {Graph Convolutional Network (GCN), Graph Attention Networks (GAT), Signed Graph Convolutional Network (SGCN), Signed Bipartite Graph Neural Networks (SBGNN) \cite{huang2021signed}, and Signed Bipartite Graph Contrastive Learning (SBCL).

\begin{itemize}[leftmargin=*]
    \item \textbf{Random Embedding:} Concatenates random 64-dimensional student ($z_{u_i} \in \mathbb{R}^{64}$) and question ($z_{v_j} \in \mathbb{R}^{64}$) embeddings, followed by logistic regression to predict.

    \item \textbf{GCN:} base on an efficient variant of convolutional neural networks which operate directly on graphs. The choice of convolutional architecture is motivated via a localized first-order approximation of spectral graph convolutions.
    
    \item \textbf{GAT:} Implemented using PyTorch Geometric (PyG) \footnote{https://pytorch-geometric.readthedocs.io/en/latest/index.html} with two convolutional layers and 300 training epochs, initializing with standard normal distribution embeddings.

    \item \textbf{SGCN:} harnesses balance theory to correctly incorporate negative links during the aggregation process. SGCN addresses the unique challenge posed by negative links, which not only carry a distinct semantic meaning compared to positive links, but also form complex relations with them due to their inherently different principles.

    \item \textbf{SBGNN:} Utilizes innovative techniques for modeling signed bipartite graphs \footnote{https://github.com/huangjunjie-cs/SBGNN}. We used publicly available code with default settings and 300 training epochs.

    \item \textbf{SBCL:} Utilizes a single-headed GAT with 2 convolutional layers and 64-dimensional random embeddings \footnote{https://github.com/Alex-Zeyu/SBGCL}. We used publicly available code with default settings and 300 training epochs. The final LLM-SBCL variant incorporates semantic question embeddings.
\end{itemize}

Data was split into training (85\%), validation (5\%), and test (10\%) sets. 64-dimensional embeddings were generated for students and questions. Question representations were enriched by concatenating stem, choices, and explanations, then processed through GPT-3.5 for knowledge points, reduced to 64 dimensions with PCA. Hyperparameters were set to $\alpha=0.8$, $\beta=5e-4$, and $r=0.1$ (where $r$ denotes the mask ratio, governing the percentage of links flipped during graph augmentation) after experimentation.

\begin{table*}[ht]
\centering
\begin{footnotesize}
\begin{tabular}{cccccccc}
\hline
 &
  \begin{tabular}[c]{@{}c@{}}Random \\ Embedding\end{tabular} &
  \multicolumn{2}{c}{\begin{tabular}[c]{@{}c@{}}Unsigned \\ Network Embedding\end{tabular}} &
  \multicolumn{2}{c}{\begin{tabular}[c]{@{}c@{}}Signed/Bipartite \\ Network Embedding\end{tabular}} &
  \multicolumn{2}{c}{\begin{tabular}[c]{@{}c@{}}Signed Bipartite graph \\ Contrastive Learning models\end{tabular}} \\ \hline
Dataset      & Random          & GCN             & GAT             & SGCN            & SBGNN           & SBCL                  & LLM-SBCL          \\ \hline
Biology      & 0.35$\pm$0.01   & 0.682$\pm$0.058 & 0.618$\pm$0.013 & 0.768$\pm$0.04  & 0.753$\pm$0.014 & { 0.772$\pm$0.016} & \textbf{0.787$\pm$0.014} \\
Law          & 0.472$\pm$0.01  & 0.823$\pm$0.01  & 0.817$\pm$0.05  & 0.84$\pm$0.013  & 0.861$\pm$0.034 & { 0.901$\pm$0.016} & \textbf{0.908$\pm$0.018} \\
Cardiff20102 & 0.136$\pm$0.062 & 0.677$\pm$0.024 & 0.571$\pm$0.013 & 0.607$\pm$0.033 & 0.712$\pm$0.016 & { 0.718$\pm$0.018} & \textbf{0.734$\pm$0.023} \\
Sydney19351  & 0.29$\pm$0.014  & 0.642$\pm$0.021 & 0.564$\pm$0.022 & 0.635$\pm$0.044 & 0.673$\pm$0.016 & { 0.674$\pm$0.021} & \textbf{0.694$\pm$0.021} \\
Sydney23146  & 0.288$\pm$0.035 & 0.728$\pm$0.013 & 0.608$\pm$0.02  & 0.726$\pm$0.04  & 0.712$\pm$0.021 & { 0.733$\pm$0.019} & \textbf{0.76$\pm$0.022}  \\ \hline
\end{tabular}
\end{footnotesize}
\caption{The results of link sign prediction on five real-world educational datasets (average binary-F1 $\pm$ standard deviation).}
\label{tab:results}
\end{table*}

\begin{table}[ht]
\centering
\begin{footnotesize}
\begin{tabular}{lc|c|c}
\hline
Dataset & Metrics & SBCL & LLM-SBCL \\ \cline{1-4}
\multirow{4}{*}{Biology}
& Binary-F1 & 0.616$\pm$0.049 & \textbf{0.629$\pm$0.061} \\
& Micro-F1 & 0.537$\pm$0.027 & \textbf{0.553$\pm$0.043} \\
& Macro-F1 & 0.512$\pm$0.015 & \textbf{0.528$\pm$0.036} \\
& AUC & 0.524$\pm$0.007 & \textbf{0.539$\pm$0.032} \\ \hline
\multirow{4}{*}{Law}
& Binary-F1 & 0.885$\pm$0.039 & \textbf{0.894$\pm$0.031} \\
& Micro-F1 & 0.803$\pm$0.059 & \textbf{0.814$\pm$0.048} \\
& Macro-F1 & \textbf{0.559$\pm$0.025} & 0.548$\pm$0.017 \\
& AUC & \textbf{0.628$\pm$0.022} & 0.594$\pm$0.017 \\ \hline
\multirow{4}{*}{Cardiff}
& Binary-F1 & 0.595$\pm$0.08 & \textbf{0.625$\pm$0.045} \\
& Micro-F1 & 0.532$\pm$0.037 & \textbf{0.543$\pm$0.029} \\
& Macro-F1 & 0.506$\pm$0.017 & \textbf{0.517$\pm$0.018} \\
& AUC & \textbf{0.521$\pm$0.019} & 0.52$\pm$0.019 \\ \hline
\multirow{4}{*}{Sydney1}
& Binary-F1 & 0.573$\pm$0.057 & \textbf{0.623$\pm$0.055} \\
& Micro-F1 & 0.539$\pm$0.035 & \textbf{0.555$\pm$0.031} \\
& Macro-F1 & 0.533$\pm$0.027 & \textbf{0.533$\pm$0.033} \\
& AUC & 0.538$\pm$0.022 & \textbf{0.544$\pm$0.026}\\ \hline
\multirow{4}{*}{Sydney2}
& Binary-F1 & 0.779$\pm$0.039 & \textbf{0.801$\pm$0.009} \\
& Micro-F1 & 0.66$\pm$0.036 & \textbf{0.681$\pm$0.012} \\
& Macro-F1 & \textbf{0.505$\pm$0.031} & 0.495$\pm$0.018 \\
& AUC & \textbf{0.518$\pm$0.017} & 0.511$\pm$0.013 \\ \hline
\end{tabular}
\end{footnotesize}
\caption{The results for cold-start problem.}
\label{tab:cold_results}
\end{table}

\subsection{Results}

In this section, \textbf{SBCL} represents our proposed model with contrastive learning, excluding NLP information. Variations of SBCL, denoted as \textbf{LLM-SBCL}, incorporate semantic embeddings from LLM. We evaluate model performance using metrics such as Area Under the Receiver Operating Characteristic Curve (AUC), binary F1, micro-F1, and macro-F1 scores. Higher metric values indicate superior link prediction.

We conducted each experiment ten times on each dataset for each model. Table \ref{tab:results} presents mean and standard deviation results for baseline models (Random Embedding, GCN, GAT, SGCN, and SBGNN) alongside our proposed SBCL models. The best results are highlighted in bold, and runner-up results are underlined. Key insights from Table \ref{tab:results} include:

% \begin{itemize}[leftmargin=*]
%     \item LLM-enhanced variants (LLM-SBCL) consistently outperform SBCL across all datasets, showcasing the benefit of incorporating question semantics.
%     \item SBCL consistently outperforms GCN and GAT, highlighting its ability to capture semantic differences (TODO XXXX: NO AUC now) between positive and negative features.
%     \item SGCN, which considers the semantic meaning of positive and negative links in signed networks, outperforms unsined methods (GCN, GAT) on most datasets.
%     \item SBCL performs better than SBGNN on the Biology dataset, except for the AUC metric (TODO XXXX: NO AUC now). This suggests that modeling student relationships, as in SBGNN, may enhance performance within learnersourcing environments like PeerWise.
% \end{itemize}

\begin{itemize}[leftmargin=*]
    \item SGCN and SBGNN, which consider the sign of the links in signed networks, outperform unsigned methods (GCN, GAT) on most datasets.
    \item SBCL consistently outperforms SGCN and SBGNN, highlighting the ability of contrastive learning.
    \item LLM-enhanced variants outperform SBCL across all datasets, showcasing the benefit of incorporating NLP semantic embeddings.
\end{itemize}

% \begin{itemize}[leftmargin=*]
%     \item Our proposed LLM-enhanced variants LLM-SBCL outperforms naive SBCL on all of the five datasets. The inclusion of question semantics does enhance the link prediction task. 
%     \item The SBCL model consistently outperforms the GCN and GAT that both are unable to capture the semantic differences between positive and negative features. The SGCN outperforms the unsined network embedding methods like GCN, GAT on most datasets, indicating the difference in semantic meaning of between positive and negative links in signed networks plays an important role in predicting students' performance.
%     %  We hypothesize this outcome could be attributed to the absence of balance degree in our dataset, as detailed in Table \ref{tab:databalance}.
%     \item On Biology dataset, SBCL shows better performance than the SBGNN except the AUC metric. This may be attributed to SBGNN's intra-set view that models student-student and question-question relationships, though not as effective as the contrastive learning in our case, appears to enhance its performance. This supports our preliminary hypothesis regarding the importance of student relationships for link prediction within learnersourcing environments like PeerWise.
% \end{itemize}

\subsection{Cold Start Problem Results}
Our study demonstrates a substantial enhancement in student performance prediction by incorporating semantic embeddings of MCQs, as shown in Table \ref{tab:cold_results}. To ensure a fair comparison, we maintained a consistent experimental setup. We randomly selected 10\% of the questions to simulate a cold start scenario. Edges related to these questions were placed in the test set, while the rest formed the training set. Our primary evaluation metric was the binary F1 score.

As a baseline, we used a generic representation learning model without semantic embeddings, which performed well with abundant data but struggled in cold start scenarios. Our enhanced model consistently outperformed the baseline across various question segments, achieving an average increase of approximately 6.4\% in binary F1 score. This improvement was particularly notable in subjects where semantic richness in MCQs was critical. The inclusion of MCQ-derived semantic embeddings holds promise for improving representation learning in e-learning environments, especially under cold start conditions. Future research can focus on optimizing these embeddings for diverse educational contexts and expanding their applicability.

\section{Conclusion and Future Work}
In this paper, we initiate the study of performance prediction for students using learnersourcing platforms. We modeled the relationship between students and questions using a signed bipartite graph. To mitigate the negative impact of noise in the datasets, which is a significant problem in learnersourcing contexts, we adopted contrastive learning as a novel learning paradigm. Additionally, we attempted to integrate both semantic and structural information related to the questions into our models. We evaluated the effectiveness of our approach on five real-world datasets from PeerWise, a widely-used learnersourcing platform. Our experimental results show that our approach achieves more accurate and stable predictions of student performance over several other baseline methods.

For future work, we aim to utilize our performance prediction model to enhance personalized learning experiences on learnersourcing platforms. Our plan involves developing a recommendation system for dynamically tailoring question suggestions based on individual student strengths and weaknesses. Exploring adaptive feedback mechanisms and real-time interventions based on predicted performance will contribute to a more effective and engaging learning environment.

\section*{Acknowledgements}
The research is supported by the Marsden Fund Council from Government funding (MFP-UOA2123), administered by the Royal Society of New Zealand.

\bibliography{aaai24}

\end{document}